\documentclass[conference]{IEEEtran}

\usepackage{cite}

\ifCLASSINFOpdf
  \usepackage[pdftex]{graphicx}
\else
  \usepackage[dvips]{graphicx}
\fi
\graphicspath{{./Figures/}}

\usepackage{amsmath,amsfonts,amssymb,stmaryrd}
\usepackage{mathtools}
\usepackage{units}
\usepackage{amsthm}

\newcommand{\norm}[1]{\left\lVert#1\right\rVert}

\DeclareMathOperator{\D}{D}

\DeclareMathOperator{\expm}{expm}
\DeclareMathOperator{\logm}{logm}
\DeclareMathOperator{\grad}{\nabla}

\DeclareMathOperator{\sym}{sym}
\DeclareMathOperator*{\Tr}{Tr}

\DeclareMathOperator*{\minimize}{minimize}
\newcommand{\NLL}{\mathcal{L}}

\newcommand{\E}{\mathbb{E}}

\newcommand*{\VEC}[1]  {\ensuremath{\boldsymbol{#1}}}
\newcommand*{\MAT}[1]  {\ensuremath{\boldsymbol{#1}}}
\newcommand{\bA}{{\MAT{A}}}
\newcommand{\bB}{\MAT{B}}
\newcommand{\bC}{\MAT{C}}
\newcommand{\bd}{\VEC{d}}
\newcommand{\bD}{\MAT{D}}

\newcommand{\bEta}{\VEC{\eta}}

\newcommand{\G}{\MAT{G}}
\newcommand{\I}{{\MAT{I}_p}}

\newcommand{\bMu}{{\VEC{\mu}}}

\newcommand{\bS}{{\MAT{S}}}
\newcommand{\bs}{{\VEC{s}}}
\newcommand{\bSigma}{{\MAT{\Sigma}}}

\newcommand{\bu}{{\VEC{u}}}

\newcommand{\bx}{{\boldsymbol{x}}}

\newcommand{\bXi}{\VEC{\xi}}

\newcommand{\bzero}{\VEC{0}}

\newcommand{\M}{{\mathcal{M}_{p, K}}}
\newcommand{\SM}{{\mathcal{SM}_{p, K}}}

\newcommand{\R}{\mathbb{R}}

\newcommand{\Spos}{{\mathcal{S}_p^{+}}}
\newcommand{\SSpos}{{\mathcal{S}\Spos}}
\newcommand{\Sym}{{\mathcal{S}_p}}

\usepackage[ruled,vlined]{algorithm2e}
\usepackage{textcomp}
\usepackage{xcolor}
\usepackage{caption}
\usepackage{subcaption}
\usepackage[hidelinks]{hyperref}

\usepackage{tikz}
\usepackage{pgfplots}
\pgfplotsset{compat=1.9}
 \pgfplotsset{every  tick label/.append style={font=\scriptsize},
 			}

\newlength\height 
\newlength\width

\begin{document}
\title{Robust Geometric Metric Learning}

\author{\IEEEauthorblockN{Antoine Collas\IEEEauthorrefmark{1}, Arnaud Breloy\IEEEauthorrefmark{2}, Guillaume Ginolhac\IEEEauthorrefmark{3}, Chengfang Ren\IEEEauthorrefmark{1}, Jean-Philippe Ovarlez\IEEEauthorrefmark{1}\IEEEauthorrefmark{4}}
	\IEEEauthorblockA{\IEEEauthorrefmark{1}SONDRA, CentraleSup\'elec, University Paris Saclay, \IEEEauthorrefmark{2}LEME, University Paris Nanterre, \\ \IEEEauthorrefmark{3}LISTIC, University Savoie Mont-Blanc, \IEEEauthorrefmark{4}DEMR, ONERA, University Paris Saclay}
}

\maketitle

\begin{abstract}
This paper proposes new algorithms for the \emph{metric learning} problem. 
We start by noticing that several classical \emph{metric learning} formulations from the literature can be viewed as modified covariance matrix estimation problems.
Leveraging this point of view, a general approach, called \emph{Robust Geometric Metric Learning} (RGML), is then studied.
This method aims at simultaneously estimating the covariance matrix of each class while shrinking them towards their (unknown) barycenter.
We focus on two specific costs functions: one associated with the Gaussian likelihood (\emph{RGML Gaussian}), and one with Tyler's $M$-estimator (\emph{RGML Tyler}).
In both, the barycenter is defined with the Riemannian distance, which enjoys nice properties of geodesic convexity and affine invariance.
The optimization is performed using the Riemannian geometry of symmetric positive definite matrices and its submanifold of unit determinant.
Finally, the performance of RGML is asserted on real datasets.
Strong performance is exhibited while being robust to mislabeled data.

\end{abstract}

\begin{IEEEkeywords}
covariance, robust estimation, Riemannian geometry, Riemannian distance, geodesic convexity, metric learning
\end{IEEEkeywords}

\section{Introduction}

Many classification algorithms rely on the distance between data points.
These algorithms include the classical \emph{K-means},  \emph{Nearest centroïd classifier}, \emph{k-nearest neighbors} and their variants.
The definition of the distance is thus of crucial importance since it determines which points will be considered similar or not, thus implies the classification rule.
In practice, classification algorithms most generally rely on the the Euclidean distance, which is $d_\I(\bx_i, \bx_j) = \norm{\bx_i - \bx_j}_2$ for $\bx_i, \bx_j \in \R^p$.
However, this distance is prone to several issues.
A pathological example is when two classes have a high variance along one common axis: within this configuration, two data points from the same class can be far away from each other, while two data points from two different classes can be very close.

To find a more relevant distance for classification, the problem of \emph{metric learning} has been proposed.
\emph{Metric learning} aims at finding a Mahalanobis distance
\begin{equation}
    d_\bA(\bx_i, \bx_j) = \sqrt{(\bx_i - \bx_j)^T \bA^{-1} (\bx_i - \bx_j)} \, ,
    \label{eq:Mahalanobis}
\end{equation}
that brings data points from same class closer, and furthers data points from different classes away.
Mathematically, \emph{metric learning} is an optimization problem of a loss function that relies on $d_\bA$.
This minimization is achieved over $\bA$, a matrix that belongs to $\Spos$ the set of $p \times p$ symmetric positive definite matrices.
The constraints of symmetricity and positivity are enforced so that $d_\bA$ is a distance.

In the following, we consider being in a supervised regime with $K$ classes, \emph{i.e.} $m$ data points $\{\bx_1, \ldots, \bx_m \}$ in $\R^p$ with their labels in $\llbracket 1, K \rrbracket$ are available.
Data points can be grouped by classes and the elements of the $k^\textup{th}$ class are denoted $\{\bx_{kl}\}$.
Then, $n_k$ pairs, $(\bx_{kl}, \bx_{kq})$ with $kl \neq kq$, of elements of the class $k$ are formed.
The set $S_k$ contains all these pairs and $S$ contains the $n_S = \sum_{k=1}^K n_k$ pairs of all the classes.
When $S$ is used, the class of a pair is not relevant, thus it is denoted by $(\bx_l, \bx_q)$ instead of $(\bx_{kl}, \bx_{kq})$.
The ratio $\frac{n_k}{n_S}$ is denoted $\pi_k$. 
Then, each vector $\bs_{ki}$ is defined as the subtraction of the elements of each pair in $S_k$, \emph{i.e.} $\bs_{ki} = \bx_{kl} - \bx_{kq}$ for $(\bx_{kl}, \bx_{kq}) \in S_k$, $i$ being the index of the pair and $l,q$ the indices of the elements of this $i^{th}$ pair.
Thus, the set $\{\bs_{ki}\}$ contains $n_k$ elements.
Then, the set $D$ contains $n_D$ pairs of vectors that do not belong to the same class.
Each vector $\bd_i$ is defined as the subtraction of the elements of each pair in $D$, \emph{i.e.} $\bd_i = \bx_l - \bx_q$ for $(\bx_l, \bx_q) \in D$.
Finally, $\Sym$ is the set of $p \times p$ symmetric matrices, $\Spos$ is the set of $p \times p$ symmetric positive definite matrices, and $\SSpos$ is the set of $p \times p$ symmetric positive definite matrices with unit determinant.

\subsection{State of the art}
Many \emph{metric learning} problems have been formulated over the years (see e.g.~\cite{survey_ML} for a complete survey).
In the following, we present notable ones that are related to our proposal.

\emph{MMC}~\cite{Xing03distancemetric} (Mahalanobis Metric for Clustering) was one of the earliest paper in this field.
This method minimizes the sum of squared distances over similar data while constraining dissimilar data to be far away from each other.
\emph{MMC} writes
\begin{equation}
    \begin{aligned}    
        \minimize_{\bA \in \Spos}& \sum_{(\bx_l, \bx_q) \in S} d_\bA^2(\bx_l, \bx_q) \\
        \textup{subject to}&  \sum_{(\bx_l, \bx_q) \in D} d_\bA(\bx_l, \bx_q) \geq 1.
    \end{aligned}
\end{equation}
Notice that $d_\bA$ (rather than $d_\bA^2$) is involved in the constraint in order to avoid a trivial rank-one solution.

Then, \emph{ITML}~\cite{Davis07} (Information-Theoretic Metric Learning) proposed to find a matrix $\bA$ that stays close to a predefined matrix $\bA_0$ while respecting constraints of similarities and dissimilarities.
The proximity between $\bA$ and $\bA_0$ is measured with the \emph{Gaussian Kullback-Leibler divergence} $\D_{KL}(\bA_0, \bA) =\Tr(\bA^{-1} \bA_0) + \log|\bA \bA_0^{-1}|$.
\emph{ITML} writes
\begin{equation}
    \begin{aligned}
        \minimize_{\bA \in \Spos}& \quad \Tr(\bA^{-1} \bA_0) + \log|\bA| \\
        \textup{subject to}&  \quad d_\bA^2(\bx_l, \bx_q) \leq u, \quad (\bx_l, \bx_q) \in S, \\
        &  \quad d_\bA^2(\bx_l, \bx_q) \geq l, \quad (\bx_l, \bx_q) \in D,
    \end{aligned}
    \label{eq:ITML_problem}
\end{equation}
where $u,v \in \R$ are threshold parameters, chosen to enforce closeness of similar points and farness of dissimilar points.
Usually $\bA_0$ is chosen as the identity matrix or as the sample covariance matrix (SCM) of the set $\{\bs_{ki}\}$.

Next, \emph{GMML} (Geometric Mean Metric Learning)~\cite{Zadeh16} is an algorithm of great interest.
Indeed, it achieves impressive performance on several datasets while being very fast thanks to a closed form formula.
The \emph{GMML} problem writes
\begin{multline}
    \!\!\!\!\!\! \minimize_{\bA \in \Spos} \frac{1}{n_S} \!\! \sum_{(\bx_l, \bx_q) \in S} \!\!\!\!\!\!\! d_\bA^2(\bx_l, \bx_q) + \frac{1}{n_D} \!\! \sum_{(\bx_l, \bx_q) \in D} \!\!\!\!\!\!\! d_{\bA^{-1}}^2(\bx_l, \bx_q).
    \label{eq:GMML_problem} 
\end{multline}
The intuition behind this problem is that $d_{\bA^{-1}}$ should be able to further away dissimilar points while $d_\bA$ close together similar points.
Then, \emph{GMML} formulation~\eqref{eq:GMML_problem} can be rewritten
\begin{equation}
    \minimize_{\bA \in \Spos} \Tr(\bA^{-1}\bS) + \Tr(\bA\bD) \, ,
    \label{eq:GMML_problem_2} 
\end{equation}
where $\bS \! = \! \frac{1}{n_S} \sum_{k=1}^K \sum_{i=1}^{n_k} \bs_{ki} \bs_{ki}^T  \textup{ and } \bD \! = \! \frac{1}{n_D} \sum_{i=1}^{n_D} \bd_i \bd_i^T$.
In~\cite{Zadeh16}, the solution of~\eqref{eq:GMML_problem_2} is derived.
It is the geodesic mid-point between $\bS^{-1}$ and $\bD$, \emph{i.e.} $\bA^{-1} = \bS^{-1} \#_{\frac12} \bD$ where
\begin{equation}
	\bS^{-1} \#_t \bD = \bS^{-\frac12} \left( \bS^{\frac12} \bD \bS^{\frac12} \right)^t \bS^{-\frac12} \textup{ with } t \in [0, 1].
    \label{eq:GMML_sol2}
\end{equation}
Then, \cite{Zadeh16} proposes to generalize this solution by $\bA^{-1} = \bS^{-1} \#_t \bD$ with $t \in [0, 1]$ (\emph{i.e.} $t$ is no longer necessarily $\frac12$).

\subsection{Metric learning as covariance matrix estimation}
\label{sec:covar_est}

In this sub-section, some \emph{metric learning} problems are expressed as covariance matrix estimation problems.

The first remark concerns the \emph{ITML} formulation~\eqref{eq:ITML_problem}.
Indeed, when the latter is written with the SCM as a prior matrix, it amounts to maximizing the likelihood of a multivariate Gaussian distribution under constraints.
Therefore, \emph{ITML} can be viewed as a \emph{covariance} matrix estimation problem.

The second remark concerns the \emph{GMML} solution of~\eqref{eq:GMML_problem_2} which is generalized to $\bA^{-1} = \bS^{-1} \#_{t} \bD$ with $t \in [0, 1]$.
In their experiments on real datasets, the authors often get their best performance with $t$ small (or even null) (see Figure 3 of~\cite{Zadeh16}).
In this case, the \emph{GMML} algorithm gives $\bA = \bS$.
This simple, yet effective, solution can be reinterpreted with an additional assumption on the data.
Let us assume that data points of each class are realizations of independent random vectors with class-dependent first and second order moments,
\begin{equation}
    \bx_{kl} \overset{d}{=} \bMu_k + \bSigma_k^{\frac12} \bu_{kl} \, ,
    \label{eq:hyp}
\end{equation}
with $\bMu_k \in \R^p$, $\bSigma_k \in \Spos$, $\E[\bu_{kl}] = \bzero$ and $\E[\bu_{kl}\bu_{kq}^T] = \I$ if $kl=kq$, $\bzero_p$ otherwise.
Thus, it follows that $\bs_{ki} \overset{d}{=} \bSigma_k^{\frac12} (\bu_{kl} - \bu_{kq})$.
Hence, the covariance matrix of $\bs_{ki}$ is twice the covariance matrix of the $k^\textup{th}$ class,
$\E[\bs_{ki}\bs_{ki}^T] \overset{d}{=} 2\bSigma_k$.
It results that, in expectation, $\bS$ is twice the arithmetic mean of the covariance matrices of the different classes,
\begin{equation}
    \E[\bS] = \frac{1}{n_S} \sum_{k=1}^K \sum_{i=1}^{n_k} \E[\bs_{ki} \bs_{ki}^T] = 2 \sum_{k=1}^K \pi_k \bSigma_k.
    \label{eq:expectation_S}
\end{equation}
The only additional assumption added to \emph{GMML} to get~\eqref{eq:expectation_S} is~\eqref{eq:hyp}.
This hypothesis is broad since it encompasses classical assumptions such as the Gaussian one.
Also notice that using $\bS$ in the Mahalanobis distance~\eqref{eq:Mahalanobis} is reminiscent of the linear discriminant analysis (LDA) pre-whitening step of the data.

\subsection{Motivations and contributions}

From Section~\ref{sec:covar_est}, \emph{GMML} can be interpreted as a $2$-steps method that computes, first, the SCM of each class and, two, their arithmetic mean.
Thus, this simple approach is not robust to outliers (\emph{e.g.} mislabeled data) since it uses the SCM as an estimator.
Moreover, other mean computation can be used, such as the Riemannian mean which benefits from many properties compared to its Euclidean counterpart~\cite{Yuan2020}.
We propose a \emph{metric learning} framework that jointly estimates regularized covariance matrices, in a robust manner, while computing their Riemannian mean.
We name this framework \emph{Riemannian Geometric Metric Learning} (\emph{RGML}).   

This idea of estimating covariance matrices while averaging them was firstly proposed in~\cite{ollila2016simultaneous}.
The novelty here is fourfold:
\textbf{1)} this formulation is applied to the problem of \emph{metric learning} (see Section~\ref{sec:RGML}),
\textbf{2)} it makes use of the Riemannian distance on $\Spos$ which was not covered by~\cite{ollila2016simultaneous} (see Section~\ref{sec:RGML}),
\textbf{3)} we leverage the Riemannian geometries of $\Spos$ and $\SSpos$~\cite{skovgaard, pennec} along with the framework of Riemannian optimization~\cite{AMS08} and hence the proposed algorithms are flexible and could be applied to other cost functions than the Gaussian and Tyler~\cite{Tyler} ones (see Section~\ref{sec:Ropt}),
\textbf{4)} the framework is applied on real datasets and shows strong performance while being robust to mislabeled data (see Section~\ref{sec:num_exp}).

\section{Problem formulation}
\label{sec:RGML}


%

\subsection{General formulation of RGML}

The formulation of the \emph{RGML} optimization problem is
\begin{equation}
    \minimize_{\substack{\theta \in \M}}  \bigg\{ h(\theta) = \sum_{k=1}^K \pi_k \left[ \NLL_k(\bA_k) + \lambda d^2(\bA, \bA_k)  \right] \bigg\} \, ,
    \label{eq:RGML}
\end{equation}
where $\theta = \left( \bA, \left\{\bA_k\right\}\right)$, $\M$ is the $K+1$ product set of $\Spos$, \emph{i.e.} $\M = \left( \Spos \right)^{K+1}$, $\NLL_k$ is a covariance matrix estimation loss on $\{\bs_{ki}\}$, $\lambda > 0$ and $d$ is a distance between matrices.
In the next subsections two costs will be considered: the Gaussian negative log-likelihood and the Tyler cost function.
Once~\eqref{eq:RGML} is achieved, the center matrix $\bA$ is used in the Mahalanobis distance~\eqref{eq:Mahalanobis} and the $\bA_k$ are discarded.
The cost function $h$ is explained more in details in  the following.

First of all, for a fixed center matrix $\bA$,~\eqref{eq:RGML} reduces to $k$ separable problems
\begin{equation}
    \minimize_{\substack{ \bA_k \in \Spos }}   \NLL_k(\bA_k) + \lambda d^2(\bA, \bA_k),
    \label{eq:RGML_fixed_A}
\end{equation}
whose solutions are estimates of $\{\bSigma_k\}$ that are regularized towards $\bA$.

Second, for $\{\bA_k\}$ fixed, solving~\eqref{eq:RGML} averages the matrices $\{\bA_k\}$.
Indeed, in this case,~\eqref{eq:RGML} reduces to
\begin{equation}
    \minimize_{\bA \in \Spos} \sum_{k=1}^K \pi_k d^2(\bA, \bA_k).
    \label{eq:RGML_fixed_A_k}
\end{equation}
For example, if $d$ is the Euclidean distance $d_E(\bA, \bA_k) = \norm{\bA - \bA_k}_2$, then the minimum of~\eqref{eq:RGML_fixed_A_k} is the arithmetic mean $\sum_{k=1}^K \pi_k \bA_k$.
In the rest of the paper, we consider the Riemannian distance on $\Spos$~\cite{skovgaard}, that is
\begin{equation}
    d_R(\bA, \bA_k) = \norm{ \logm\left(\bA^{-\frac12} \bA_k \bA^{-\frac12} \right) }_2
    \label{eq:dist_Riemannian}
\end{equation}
with $\logm$ being the matrix logarithm.
A nice property of $d_R$~\eqref{eq:dist_Riemannian} is its affine invariance.
Indeed, for any $\bC$ invertible, we have $d_R(\bC \bA \bC^T, \bC \bA_k \bC^T) = d_R(\bA, \bA_k)$.
Thus, if $\left\{\bs_{ki}\right\}$ is transformed to $\left\{\bC \bs_{ki}\right\}$ then the minimum $\left(\bA, \left\{\bA_k\right\}\right)$ of~\eqref{eq:RGML_2} becomes $\left(\bC\bA\bC^T, \left\{\bC\bA_k\bC^T\right\}\right)$.
Another nice property of this distance is its geodesic convexity, as it will be discussed in Section~\ref{sec:Ropt}.

With this Riemannian distance, the general formulation of the \emph{RGML} optimization problem~\eqref{eq:RGML} becomes
\begin{equation}
    \minimize_{\substack{\theta \in \M}}  \bigg\{ \! h(\theta) = \sum_{k=1}^K \! \pi_k \! \left[ \NLL_k(\bA_k) + \lambda d_R^2(\bA, \bA_k)  \right] \bigg\}.
    \label{eq:RGML_2}
\end{equation}
We emphasis that the optimization of~\eqref{eq:RGML_2} is performed with respect to all the matrices $\bA$ and $\{\bA_k\}$ at the same time.
Thus it both estimates regularized covariance matrices $\{\bA_k\}$ while averaging them to estimate their unknown barycenter $\bA$.

\subsection{RGML Gaussian}


To get a practical cost function $h$~\eqref{eq:RGML_2}, it only remains to specify the functions $\NLL_k$.
The most classical assumption on the data distribution is the Gaussian one (\emph{e.g.} considered in \emph{ITML} with the SCM as prior).
Thus, the first functions $\NLL_k$ considered are the centered multivariate Gaussian negative log-likelihoods
\begin{equation}
    \NLL_{G,k}(\bA) = \frac{1}{n_k} \sum_{i=1}^{n_k} \bs_{ki}^T \bA^{-1} \bs_{ki} + \log |\bA|.
\end{equation}
With this negative log-likelihood, the \emph{RGML} optimization problem~\eqref{eq:RGML_2} becomes
\begin{equation}
    \minimize_{\substack{\theta \in \M}} \! \bigg\{ \! h_G(\theta) \! = \! \sum_{k=1}^K \! \pi_k \! \left[ \NLL_{G,k}(\bA_k) \! + \! \lambda d_R^2(\bA, \bA_k)  \right] \!\! \bigg\}\!\!.\!\!
    \label{eq:h_G}
\end{equation}

\subsection{RGML Tyler}

When data is non-Gaussian, robust covariance matrix estimation methods are a preferred choice.
This occurs whenever the probability distribution of the data is heavy-tailed or a small proportion of the samples represents outlier behavior.
In a classification setting, the latter happens when data are mislabeled.
A classical robust estimator is the Tyler's estimator~\cite{Tyler} which is the minimizer of the following cost function
\begin{equation}
    \NLL_{T,k}(\bA) = \frac{p}{n_k} \sum_{i=1}^{n_k} \log\left(\bs_{ki}^T \bA^{-1} \bs_{ki}\right) + \log |\bA|.
    \label{eq:Tyler_cost}
\end{equation}
An important remark is that~\eqref{eq:Tyler_cost} is invariant to the scale of $\bA$.
Indeed $\forall \alpha > 0$, it is easily checked that $\NLL_{T,k}(\alpha \bA) = \NLL_{T,k}(\bA)$.
Thus, a constraint of unit determinant is added to~\eqref{eq:RGML_2} to fix the scales of $\{\bA_k\}$.
Furthermore, the Riemannian distance~\eqref{eq:dist_Riemannian} is also the one on $\SSpos$.
Thus, we choose to also constrain $\bA$ so that it is the Riemannian mean of $\{\bA_k\}$ on $\SSpos$.
We denote by $\SM$ this new parameter space
\begin{equation}
    \SM \! = \! \left\{ \theta \! \in \! \M, |\bA| = |\bA_k| = 1,  \, \forall k \in \llbracket 1, K \rrbracket \right\}.
    \label{eq:SM}
\end{equation}
Thus, the \emph{RGML} optimization problem~\eqref{eq:RGML_2} with the Tyler cost function~\eqref{eq:Tyler_cost} becomes
\begin{equation}
    \minimize_{\substack{\theta \in \SM}}  \! \bigg\{ \! h_{T}(\theta) \! = \! \sum_{k=1}^K \! \pi_k \! \left[ \NLL_{T,k}(\bA_k) \! + \! \lambda d_R^2(\bA, \bA_k)  \right] \!\! \bigg\} \!\!.\!\!
    \label{eq:h_T}
\end{equation}

\section{Riemannian optimization}
\label{sec:Ropt}

The objective of this section is to present the Algorithms~\ref{algo:RMLGaussien} and~\ref{algo:RMLTyler} which minimize~\eqref{eq:h_G} and~\eqref{eq:h_T} respectively.
They leverage the Riemannian optimization framework~\cite{AMS08, boumal2022intromanifolds}.
The products manifolds $\M$ and $\SM$ (directly inherited from $\Spos$ and $\SSpos$~\cite{skovgaard, pennec}) are presented.

\subsection{Riemannian optimization and g-convexity on $\M$}

Since, $\M$ is an open set in a vector space, the tangent space $T_\theta\M$ (linearization of the Riemannian manifold at a given point)  is identified to $\left(\Sym\right)^{K+1}$.
Then, the affine invariant metric is chosen as the Riemannian metric~\cite{skovgaard},
$\forall \xi = \left(\bXi, \left\{\bXi_k\right\}\right), \forall \eta = \left(\bEta, \left\{\bEta_k\right\} \right) \in T_\theta\M$
\begin{equation}
    \langle \xi, \eta \rangle_\theta^\M  \!\! = \!\! \Tr \! \left( \bA^{-1} \bXi \bA^{-1} \bEta \right) + \sum_{k=1}^K \! \Tr \! \left( \bA_k^{-1} \bXi_k \bA_k^{-1} \bEta_k \right). \!\!\!
    \label{eq:FIM}
\end{equation}
Thus the orthogonal projection from the ambient space onto the tangent space at $\theta$ is
\begin{equation}
    P_\theta^\M(\xi) = \left( \sym(\bXi), \left\{\sym(\bXi_k)\right\}\right) \, ,
\end{equation}
where $\sym(\bXi) = \frac12 (\bXi + \bXi^T )$.
Then, the exponential map (function that maps tangent vectors, such as gradients of loss functions, to points on the manifold) is 
\begin{equation}
    \exp_\theta^\M (\xi) = \Big( \exp_\bA^\Spos (\bXi), \Big\{\exp_{\bA_k}^\Spos (\bXi_k)\Big\} \Big) \, ,
    \label{eq:exp_map}
\end{equation}
where $\exp_\bA^\Spos (\bXi) = \bA \expm(\bA^{-1} \bXi)$ with $\expm$ being the matrix exponential.
Then, for a loss function $\ell: \M \rightarrow \R$, the Riemannian gradient at $\theta$ denoted $\grad_\M \ell(\theta)$ is defined as the unique element such that $\forall \xi \in T_\theta\M$, $\D \ell(\theta)[\xi] = \langle \grad_\M \ell(\theta), \xi \rangle_\theta^\M$ where $\D$ is the directional derivative.
It results that
\begin{equation}
    \grad_\M \ell(\theta) = P_\theta^\M \left(\bA \G \bA, \{\bA_k \G_k \bA_k\}\right) \, ,
    \label{eq:grad_M}
\end{equation}
where $\left(\G, \left\{\G_k\right\}\right)$ is the classical Euclidean gradient of $\ell$ at $\theta$.
In practice this Euclidean gradient can be computed using automatic differentiation libraries such as JAX~\cite{jax}.
With the exponential map~\eqref{eq:exp_map}, and the Riemannian gradient~\eqref{eq:grad_M}, we have the main tools to minimize~\eqref{eq:h_G}.
However, to improve the numerical stability, a retraction (approximation of the exponential map~\eqref{eq:exp_map}) is preferred,
\begin{equation}
    R_\theta^\M (\xi) = \Big( R_\bA^\Spos(\bXi), \Big\{R_{\bA_k}^\Spos(\bXi_k)\Big\} \Big) \, ,
\end{equation}
where $R_\bA^\Spos (\bXi) = \bA + \bXi + \frac12 \bXi \bA^{-1} \bXi$.
A Riemannian gradient descent minimizing~\eqref{eq:h_G} is presented in Algorithm~\ref{algo:RMLGaussien}. 
\begin{algorithm}[t]
    \KwIn{Data $\{\bs_{ki}\}$, initialization $\theta^{(0)} \in \M$}
    \KwOut{$\theta^{(t)} \in \M$}
    \For{$t=0$ \textbf{to convergence}}{
        Compute a step size $\alpha$ (see \cite[Ch. 4]{AMS08}) and set
	    $\theta^{(t+1)} = R_{\theta^{(t)}}^\M\left(-\alpha \grad_\M h_G(\theta^{(t)})\right)$
    }
    \caption{Riemannian gradient descent to minimize $h_G$~\eqref{eq:h_G}}
    \label{algo:RMLGaussien}
\end{algorithm}

We finish this subsection by presenting the geodesic convexity of $h_G$~\eqref{eq:h_G} on $\M$ (see~\cite[Chapter 11]{boumal2022intromanifolds} for a presentation of the geodesic convexity).
First of all, the geodesic on $\M$ between $a = \left(\bA, \left\{\bA_k\right\}\right)$ and $b = \left(\bB, \left\{\bB_k\right\}\right)$ is
\begin{equation}
    a \#_t b \! = \! \left( \bA \#_t \bB, \left\{\bA_k \#_t \bB_k\right\} \right) \, ,
    \label{eq:geodesic_M}
\end{equation}
where $\#$ is the geodesic~\eqref{eq:GMML_sol2} on $\Spos$ and $t \in [0, 1]$.
Then, a loss function $\ell$ is said to be geodesically convex (or g-convex) if
\begin{equation}
    \ell \left(a \#_t b\right) \leq t \, \ell(a) + (1-t) \, \ell(b), \quad \forall t \in [0, 1].
\end{equation}
If $\ell$ is g-convex, then any local minimizer is a global minimizer.
\cite{ollila2016simultaneous} proves that $h_G$~\eqref{eq:h_G} is g-convex.
Hence, any local minimizer of~\eqref{eq:h_G} is a global minimizer.

\subsection{$\SM$: a geodesic submanifold of $\M$}
In~\eqref{eq:SM}, $\SM$ is defined as a subset of $\M$.
In fact, $\SM$ can even be turned into a Riemannian submanifold of $\M$.
First of all, the tangent space of $\SM$ at $\theta$ is
\begin{multline}
    T_\theta \SM = \Big\{ \xi \in T_\theta \M: \Tr(\bA^{-1} \bXi) = 0, \\ \Tr(\bA_k^{-1} \bXi_k) = 0 \quad \forall k \in \llbracket 1, K \rrbracket \Big\}.
\end{multline}
By endowing $\SM$ with the Riemannian metric of $\M$, it becomes a Riemannian submanifold.
$\forall \xi, \eta \in T_\theta\SM$ we have $\langle \xi,\eta \rangle_\theta^\SM = \langle \xi,\eta \rangle_\theta^\M$.
The orthogonal projection from the ambient space onto the tangent space at $\theta$ is
\begin{equation}
    P_\theta^\SM(\xi) = \Big( P_\bA^\SSpos (\bXi), \Big\{P_{\bA_k}^\SSpos (\bXi_k)\Big\}\Big) \, ,
\end{equation}
where $P_\bA^\SSpos (\bXi) = \sym\left(\bXi\right) - \frac{1}{p} \Tr\left( \bA^{-1} \sym\left(\bXi\right)\right) \bA$.
A remarkable result is that $\SM$ is a geodesic submanifold of $\M$, \emph{i.e.}, the geodesics of $\SM$ are those of $\M$.
It results that the exponential mapping on $\SM$ is $\exp_\theta^\SM(\xi) = \exp_\theta^\M(\xi)$.
Then, for a loss function $\ell: \SM \rightarrow \R$, the Riemannian gradient at $\theta$ is
\begin{equation}
    \grad_\SM \ell(\theta) = P_\theta^\SM \left(\bA \G \bA, \{\bA_k \G_k \bA_k\}\right) \, ,
    \label{eq:grad_SM}
\end{equation}
where $\left(\G, \left\{\G_k\right\}\right)$ is the classical Euclidean gradient of $\ell$ at $\theta$.
Once again, a retraction that approximates the exponential mapping is leveraged to improve the numerical stability,
\begin{equation}
    R_\theta^\SM (\xi) = \Big( R_\bA^\SSpos(\bXi), \Big\{ R_{\bA_k}^\SSpos(\bXi_k) \Big\} \Big) \, ,
\end{equation}
where $R_\bA^\SSpos (\bXi) = \displaystyle \frac{\bA + \bXi + \frac12 \bXi \bA^{-1} \bXi}{\left| \bA + \bXi + \frac12 \bXi \bA^{-1} \bXi \right|^{\frac{1}{p}}}$.

Finally, $h_T$~\eqref{eq:h_T} is g-convex on $\SM$.
Indeed, \cite{ollila2016simultaneous} proved that $h_T$ is g-convex on $\M$ and $\SM$ is a geodesic submanifold of $\M$.

\begin{algorithm}[t]
    \KwIn{Data $\{\bs_{ki}\}$, initialization $\theta^{(0)} \in \SM$}
    \KwOut{$\theta^{(t)} \in \SM$}
    \For{$t=0$ \textbf{to convergence}}{
        Compute a step size $\alpha$ (see \cite[Ch. 4]{AMS08}) and set
	    $\theta^{(t+1)} \! = \! R_{\theta^{(t)}}^\SM\left(-\alpha \grad_\SM h_T(\theta^{(t)})\right)$
    }
    \caption{Riemannian gradient descent to minimize $h_T$~\eqref{eq:h_T}}
    \label{algo:RMLTyler}
\end{algorithm}

\section{Experiments}
\label{sec:num_exp}

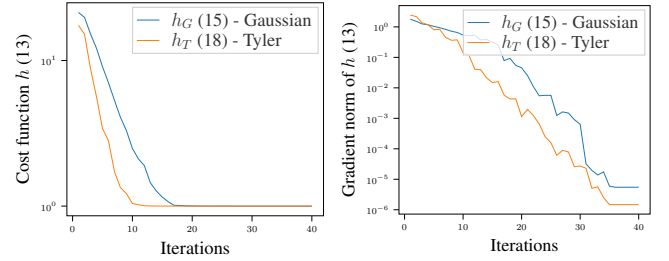
\begin{figure}[h]
    \centering
    \begin{subfigure}{.49\linewidth}
        \resizebox{\linewidth}{!}{
\begin{tikzpicture}

\definecolor{darkgray176}{RGB}{176,176,176}
\definecolor{darkorange25512714}{RGB}{255,127,14}
\definecolor{lightgray204}{RGB}{204,204,204}
\definecolor{steelblue31119180}{RGB}{31,119,180}

\begin{axis}[
legend cell align={left},
legend style={fill opacity=0.8, draw opacity=1, text opacity=1, draw=lightgray204},
log basis y={10},
tick align=outside,
tick pos=left,
x grid style={darkgray176},
xlabel={\Large Iterations},
xmin=-0.95, xmax=41.95,
xtick style={color=black},
y grid style={darkgray176},
ylabel={\Large Cost function $h$~\eqref{eq:RGML_2}},
ymin=0.858012257492904, ymax=24.9250859192437,
ymode=log,
ytick style={color=black},
ytick={0.01,0.1,1,10,100,1000},
yticklabels={
  \(\displaystyle {10^{-2}}\),
  \(\displaystyle {10^{-1}}\),
  \(\displaystyle {10^{0}}\),
  \(\displaystyle {10^{1}}\),
  \(\displaystyle {10^{2}}\),
  \(\displaystyle {10^{3}}\)
}
]
\addplot [semithick, steelblue31119180]
table {%
1 21.3860292377749
2 19.7624682825012
3 15.2064758867148
4 12.130294249295
5 9.07082878001858
6 7.03808290328894
7 5.33090191809333
8 4.0751668756779
9 3.29968172427439
10 2.48819966841609
11 2.11045937629324
12 1.89573759019214
13 1.44573070455362
14 1.26904724098023
15 1.15327401122611
16 1.07036833537763
17 1.0118844302825
18 1.0100892880357
19 1.00441236889389
20 1.00276261038328
21 1.00094943034069
22 1.00028824862627
23 1.0001233234711
24 1.00005631737005
25 1.00004130678564
26 1.00000787519909
27 1.00000271127043
28 1.00000226797137
29 1.00000084230669
30 1.00000041497145
31 1.00000000232295
32 1.00000000100053
33 1.00000000051951
34 1.00000000040882
35 1.0000000001156
36 1
37 1
38 1
39 1
40 1
};
\addlegendentry{\Large $h_G$~\eqref{eq:h_G} - Gaussian}
\addplot [semithick, darkorange25512714]
table {%
1 17.4187330424892
2 15.1425282993392
3 8.78172325381668
4 5.62961502382682
5 3.4119053836599
6 2.81267109844388
7 1.72997965121175
8 1.33813397401285
9 1.21957249076698
10 1.0421156216582
11 1.02225817090194
12 1.00538224207378
13 1.0027393794889
14 1.00098850084568
15 1.00044953575357
16 1.0003499908054
17 1.00008397593522
18 1.00004002323224
19 1.00002770563135
20 1.00000525823397
21 1.00000446443536
22 1.00000187614526
23 1.000000571092
24 1.00000011829799
25 1.00000005084043
26 1.00000001759529
27 1.00000000940989
28 1.00000000710071
29 1.00000000098339
30 1.00000000091772
31 1.00000000067634
32 1.00000000013515
33 1.00000000012398
34 1.00000000004155
35 1
36 1
37 1
38 1
39 1
40 1
};
\addlegendentry{\Large $h_T$~\eqref{eq:h_T} - Tyler}
\end{axis}

\end{tikzpicture}
        }
    \end{subfigure}%
    \begin{subfigure}{.49\linewidth}
        \resizebox{\linewidth}{!}{
\begin{tikzpicture}

\definecolor{darkgray176}{RGB}{176,176,176}
\definecolor{darkorange25512714}{RGB}{255,127,14}
\definecolor{lightgray204}{RGB}{204,204,204}
\definecolor{steelblue31119180}{RGB}{31,119,180}

\begin{axis}[
legend cell align={left},
legend style={fill opacity=0.8, draw opacity=1, text opacity=1, draw=lightgray204},
log basis y={10},
tick align=outside,
tick pos=left,
x grid style={darkgray176},
xlabel={\Large Iterations},
xmin=-0.95, xmax=41.95,
xtick style={color=black},
y grid style={darkgray176},
ylabel={\Large Gradient norm of $h$~\eqref{eq:RGML_2}},
ymin=7.21422884052217e-07, ymax=4.98603298518297,
ymode=log,
ytick style={color=black},
ytick={1e-08,1e-07,1e-06,1e-05,0.0001,0.001,0.01,0.1,1,10,100},
yticklabels={
  \(\displaystyle {10^{-8}}\),
  \(\displaystyle {10^{-7}}\),
  \(\displaystyle {10^{-6}}\),
  \(\displaystyle {10^{-5}}\),
  \(\displaystyle {10^{-4}}\),
  \(\displaystyle {10^{-3}}\),
  \(\displaystyle {10^{-2}}\),
  \(\displaystyle {10^{-1}}\),
  \(\displaystyle {10^{0}}\),
  \(\displaystyle {10^{1}}\),
  \(\displaystyle {10^{2}}\)
}
]
\addplot [semithick, steelblue31119180]
table {%
1 1.78876483093685
2 1.52586434453736
3 1.25014346184431
4 1.17292711612362
5 1.04161947433283
6 0.939335381538934
7 0.828971465870026
8 0.720041698306668
9 0.655306346373388
10 0.538048377571707
11 0.515254930664259
12 0.529939039932821
13 0.372841215339365
14 0.388292778098904
15 0.31498836379574
16 0.254241931835094
17 0.0786004071214253
18 0.0925444106351285
19 0.0544870095671396
20 0.0452351748857362
21 0.0254083621273166
22 0.0110186082542974
23 0.00549704386396503
24 0.00568783094083476
25 0.00565405738458728
26 0.00124122511199817
27 0.00162439797857274
28 0.00149117219043474
29 0.000906891603320112
30 0.00063726870519171
31 3.20335071730067e-05
32 1.95597257101154e-05
33 1.3902272412574e-05
34 1.74605000602177e-05
35 5.94408923254347e-06
36 5.48312717922588e-06
37 5.48312717922588e-06
38 5.48312717922588e-06
39 5.48312717922588e-06
40 5.48312717922588e-06
};
\addlegendentry{\Large $h_G$~\eqref{eq:h_G} - Gaussian}
\addplot [semithick, darkorange25512714]
table {%
1 2.43705773809125
2 2.17525594793763
3 1.35909884519224
4 1.17658073898635
5 0.803590796787558
6 0.83508055339157
7 0.453446364983733
8 0.363951642557338
9 0.374799687534714
10 0.141716232245602
11 0.117727312824445
12 0.0406916757049817
13 0.0394950931343707
14 0.0213396973803747
15 0.0148987979361156
16 0.0159774732468674
17 0.00574350403371805
18 0.00432310717417313
19 0.00436417475241643
20 0.00114082021569172
21 0.00195840453152191
22 0.00123609637222978
23 0.000659754067924992
24 0.000251325108474552
25 0.000159496348661114
26 6.13900903569315e-05
27 8.88571983942491e-05
28 7.86330863445931e-05
29 2.58771449765248e-05
30 2.72953541203615e-05
31 2.29938442481343e-05
32 5.0543352483572e-06
33 5.6544000886604e-06
34 2.50849055623794e-06
35 1.4759758211422e-06
36 1.4759758211422e-06
37 1.4759758211422e-06
38 1.4759758211422e-06
39 1.4759758211422e-06
40 1.4759758211422e-06
};
\addlegendentry{\Large $h_T$~\eqref{eq:h_T} - Tyler}
\end{axis}

\end{tikzpicture}
        }
    \end{subfigure}%
    \caption{
    Left: Gaussian~\eqref{eq:h_G} and Tyler~\eqref{eq:h_T} costs functions with respect to the number of iterations of Algorithms~\ref{algo:RMLGaussien} and~\ref{algo:RMLTyler} respectively.
    Right: Riemannian gradient norms of Gaussian~\eqref{eq:h_G} and Tyler~\eqref{eq:h_T} costs functions.
    The optimization is performed on the \emph{Wine} dataset.
    }
    \label{fig:iter}
\end{figure}

\setlength{\tabcolsep}{8pt}
\begin{table*}[t]
    \center
    \begin{tabular}{ |c|c c c c|c c c c|c c c c| } 
        \hline
        & \multicolumn{4}{|c|}{Wine} & \multicolumn{4}{|c|}{Vehicle} & \multicolumn{4}{|c|}{Iris}  \\
        & \multicolumn{4}{|c|}{$p=13$ , $n=178$, $K=3$} & \multicolumn{4}{|c|}{$p=18$, $n=846$, $K=4$} & \multicolumn{4}{|c|}{$p=4$, $n=150$, $K=3$}  \\
        \hline
        Method & \multicolumn{4}{c|}{Mislabeling rate} & \multicolumn{4}{c|}{Mislabeling rate} &  \multicolumn{4}{c|}{Mislabeling rate} \\
        & 0$\%$ & 5$\%$ & 10$\%$ & 15$\%$ & 0$\%$ & 5$\%$ & 10$\%$ & 15$\%$ & 0$\%$ & 5$\%$ & 10$\%$ & 15$\%$ \\
        \hline
        Euclidean & 30.12 & 30.40 & 31.40 & 32.40 & 38.27 & 38.58 & 39.46 & 40.35 & 3.93 & 4.47 & 5.31 & \textbf{6.70} \\ 
        SCM & 10.03 & 11.62 & 13.70 & 17.57 & 23.59 & 24.27 & 25.24 & 26.51 & 12.57 & 13.38 & 14.93 & 16.68 \\ 
        ITML - Identity & 3.12 & 4.15 & 5.40 & \textbf{7.74} & 24.21 & 23.91 & 24.77 & 26.03 & 3.04 & 4.47 & 5.31 & \textbf{6.70} \\ 
        ITML - SCM & 2.45 & 4.76 & 6.71 & 10.25 & 23.86 & 23.82 & 24.89 & 26.30 & 3.05 & 13.38 & 14.92 & 16.67 \\ 
        GMML & 2.16 & 3.58 & 5.71 & 9.86 & 21.43 & 22.49 & 23.58 & 25.11 & 2.60 & 5.61 & 9.30 & 12.62 \\ 
        LMNN & 4.27 & 6.47 & 7.83 & 9.86 & 20.96 & 24.23 & 26.28 & 28.89 & 3.53 & 9.59 & 11.19 & 12.22 \\ 
        \hline
        RGML - Gaussian & \textbf{2.07} & \textbf{2.93} & 5.15 & 9.20 & \textbf{19.76} & 21.19 & 22.52 & 24.21 & \textbf{2.47} & 5.10 & 8.90 & 12.73 \\ 
        RGML - Tyler & \textbf{2.12} & \textbf{2.90} & \textbf{4.51} & 8.31 & 19.90 & \textbf{20.96} & \textbf{22.11} & \textbf{23.58} & \textbf{2.48} & \textbf{2.96} & \textbf{4.65} & 7.83 \\ 
        \hline
    \end{tabular}
    \caption{
    Misclassification errors on 3 datasets: Wine, Vehicle and Iris.
    Best results and those within $0.05\%$ are in \textbf{bold}.
    The mislabeling rates indicate the percentage of labels that are randomly changed in the training set.
    }
    \label{table:results}
\end{table*}

In this section, we exhibit a practical interest of the \emph{RGML} method developed in Sections~\ref{sec:RGML} and~\ref{sec:Ropt}.
All implementations of the following experiments are available at~\url{https://github.com/antoinecollas/robust_metric_learning}.
We apply it on real datasets from the \emph{UCI machine learning repository}~\cite{UCI}.
The three considered datasets are: \emph{Wine}, \emph{Vehicle}, and \emph{Iris}.
They are classification datasets, and their data dimensions along with their number of classes are presented in Table~\ref{table:results}.
These datasets are well balanced, \emph{i.e.} they roughly have the same number of data for all the classes.
The numbers of generated pairs in $S$ and $D$ are $n_S = n_D = 75\,K(K-1)$ (as in \cite{Davis07} and \cite{Zadeh16}).

The classification is done following a very classical protocol in \emph{metric learning}.
\textbf{1)} A matrix $\bA$ is estimated via a \emph{metric learning} method.
\textbf{2)} The data $\{\bx_l\}$ are multiplied by $\bA^{-\frac{1}{2}}$ to get $\{\bA^{-\frac{1}{2}} \bx_l\}$.
\textbf{3)} The data $\{\bA^{-\frac{1}{2}} \bx_l\}$ are classified using a \emph{k-nearest neighbors} with $5$ neighbors.
Thus, the classification is performed using the Mahalanobis distance $d_\bA$ defined by \eqref{eq:Mahalanobis} in the Introduction.
This classification is repeated $200$ times via cross-validation.
The proportion of the training/test sets is $50/50$.
The error of classification is computed for each fold and the mean error is reported in Table~\ref{table:results}.
In order to show the robustness of the proposed method, mislabeled data are introduced.
To do so, we randomly select data in the training set whose labels are then randomly changed for new labels.

The implementations of the cross-validation as well as the \emph{k-nearest neighbors} are from the scikit-learn library~\cite{scikit-learn}.
The proposed methods \emph{RGML Gaussian} and \emph{RGML Tyler} have been implemented using JAX~\cite{jax}.
The chosen value of parameter $\lambda$ is $0.05$.
Its value has little impact on performance as long as it is neither too small nor too large.
The proposed algorithms are compared to the classical \emph{metric learning} algorithms:
the identity matrix (called Euclidean in Table~\ref{table:results}),
the SCM computed on all the data,
\emph{ITML}~\cite{Davis07},
\emph{GMML}~\cite{Zadeh16},
and  \emph{LMNN}~\cite{Weinberger09}.
The implementations of the metric-learn library~\cite{metric-learn} are used for the last three algorithms.

From Table~\ref{table:results}, several observations are made.
First of all, on the raw data (\emph{i.e.} when the mislabeling rate is $0\%$) the \emph{RGML Gaussian} is always the best performing algorithm among those tested.
Also, the \emph{RGML Tyler} always comes close with a maximum discrepancy of $0.26\%$ versus the \emph{RGML Gaussian}.
Then, the \emph{RGML Tyler} is the best performing algorithm when the mislabeling rate is $5\%$ or $10\%$.
When the mislabeling rate is $15\%$, \emph{RGML Tyler} is the best performing algorithm for the \emph{Vehicle} dataset and it is only beaten by \emph{ITML - Identity} on the two other datasets.
This shows the interest of considering robust cost functions such the Tyler's cost function~\eqref{eq:Tyler_cost} in the presence of poor labeling.

Finally, the \emph{RGML} algorithms are fast.
Indeed, Figure~\ref{fig:iter} shows that both \emph{RGML Gaussian} and \emph{RGML Tyler} converge in less than $20$ iterations on the \emph{Wine} dataset.

\section{Conclusions}
This paper has proposed to view some classical \emph{metric learning} problems as covariance matrix estimation problems.
From this point of view, the \emph{RGML} optimization problem has been formalized.
It aims at estimating regularized covariance matrices, in a robust manner, while computing their Riemannian mean.
The formulation is broad and several more specific costs functions have been studied.
The first one leverages the classical Gaussian likelihood and the second one the Tyler's cost function.
In both cases, the \emph{RGML} problem is g-convex and thus any local minimizer is a global one.
Two Riemannian-based optimization algorithms are proposed to minimize these cost functions.
Finally, the performance of the proposed approach is studied on several datasets.
They improve the classification accuracy and are robust to mislabeled data.

\bibliographystyle{IEEEtran}
\bibliography{references}

\begin{thebibliography}{10}
\providecommand{\url}[1]{#1}
\csname url@samestyle\endcsname
\providecommand{\newblock}{\relax}
\providecommand{\bibinfo}[2]{#2}
\providecommand{\BIBentrySTDinterwordspacing}{\spaceskip=0pt\relax}
\providecommand{\BIBentryALTinterwordstretchfactor}{4}
\providecommand{\BIBentryALTinterwordspacing}{\spaceskip=\fontdimen2\font plus
\BIBentryALTinterwordstretchfactor\fontdimen3\font minus
  \fontdimen4\font\relax}
\providecommand{\BIBforeignlanguage}[2]{{%
\expandafter\ifx\csname l@#1\endcsname\relax
\typeout{** WARNING: IEEEtran.bst: No hyphenation pattern has been}%
\typeout{** loaded for the language `#1'. Using the pattern for}%
\typeout{** the default language instead.}%
\else
\language=\csname l@#1\endcsname
\fi
#2}}
\providecommand{\BIBdecl}{\relax}
\BIBdecl

\bibitem{survey_ML}
\BIBentryALTinterwordspacing
J.~L. Suárez, S.~García, and F.~Herrera, ``A tutorial on distance metric
  learning: Mathematical foundations, algorithms, experimental analysis,
  prospects and challenges,'' \emph{Neurocomputing}, vol. 425, pp. 300--322,
  2021. [Online]. Available:
  \url{https://www.sciencedirect.com/science/article/pii/S0925231220312777}
\BIBentrySTDinterwordspacing

\bibitem{Xing03distancemetric}
E.~P. Xing, A.~Y. Ng, M.~I. Jordan, and S.~Russell, ``Distance metric learning,
  with application to clustering with side-information,'' in \emph{Advances in
  Neural Information Processing Systems 15}.\hskip 1em plus 0.5em minus
  0.4em\relax MIT Press, 2003, pp. 505--512.

\bibitem{Davis07}
\BIBentryALTinterwordspacing
J.~V. Davis, B.~Kulis, P.~Jain, S.~Sra, and I.~S. Dhillon,
  ``Information-theoretic metric learning,'' in \emph{Proceedings of the 24th
  International Conference on Machine Learning}, ser. ICML '07.\hskip 1em plus
  0.5em minus 0.4em\relax New York, NY, USA: Association for Computing
  Machinery, 2007, p. 209–216. [Online]. Available:
  \url{https://doi.org/10.1145/1273496.1273523}
\BIBentrySTDinterwordspacing

\bibitem{Zadeh16}
P.~H. Zadeh, R.~Hosseini, and S.~Sra, ``Geometric mean metric learning,'' in
  \emph{Proceedings of the 33rd International Conference on International
  Conference on Machine Learning - Volume 48}, ser. ICML'16.\hskip 1em plus
  0.5em minus 0.4em\relax JMLR.org, 2016, p. 2464–2471.

\bibitem{Yuan2020}
\BIBentryALTinterwordspacing
X.~Yuan, W.~Huang, P.-A. Absil, and K.~A. Gallivan, \emph{Averaging Symmetric
  Positive-Definite Matrices}.\hskip 1em plus 0.5em minus 0.4em\relax Cham:
  Springer International Publishing, 2020, pp. 555--575. [Online]. Available:
  \url{https://doi.org/10.1007/978-3-030-31351-7\_20}
\BIBentrySTDinterwordspacing

\bibitem{ollila2016simultaneous}
\BIBentryALTinterwordspacing
E.~Ollila, I.~Soloveychik, D.~E. Tyler, and A.~Wiesel, ``Simultaneous penalized
  m-estimation of covariance matrices using geodesically convex optimization,''
  2016. [Online]. Available: \url{https://arxiv.org/abs/1608.08126}
\BIBentrySTDinterwordspacing

\bibitem{skovgaard}
\BIBentryALTinterwordspacing
L.~T. Skovgaard, ``A riemannian geometry of the multivariate normal model,''
  \emph{Scandinavian Journal of Statistics}, vol.~11, no.~4, pp. 211--223,
  1984. [Online]. Available: \url{http://www.jstor.org/stable/4615960}
\BIBentrySTDinterwordspacing

\bibitem{pennec}
X.~Pennec, P.~Fillard, and N.~Ayache, ``A riemannian framework for tensor
  computing,'' \emph{International Journal of Computer Vision}, vol.~66, pp.
  41--66, 2005.

\bibitem{AMS08}
P.-A. Absil, R.~Mahony, and R.~Sepulchre, \emph{Optimization algorithms on
  matrix manifolds}.\hskip 1em plus 0.5em minus 0.4em\relax Princeton
  University Press, 2008.

\bibitem{Tyler}
\BIBentryALTinterwordspacing
D.~E. Tyler, ``{A Distribution-Free $M$-Estimator of Multivariate Scatter},''
  \emph{The Annals of Statistics}, vol.~15, no.~1, pp. 234 -- 251, 1987.
  [Online]. Available: \url{https://doi.org/10.1214/aos/1176350263}
\BIBentrySTDinterwordspacing

\bibitem{boumal2022intromanifolds}
\BIBentryALTinterwordspacing
N.~Boumal, ``An introduction to optimization on smooth manifolds,'' To appear
  with Cambridge University Press, Jan 2022. [Online]. Available:
  \url{http://www.nicolasboumal.net/book}
\BIBentrySTDinterwordspacing

\bibitem{jax}
\BIBentryALTinterwordspacing
J.~Bradbury, R.~Frostig, P.~Hawkins, M.~J. Johnson, C.~Leary, D.~Maclaurin,
  G.~Necula, A.~Paszke, J.~Vander{P}las, S.~Wanderman-{M}ilne, and Q.~Zhang,
  ``{JAX}: composable transformations of {P}ython+{N}um{P}y programs,'' 2018.
  [Online]. Available: \url{http://github.com/google/jax}
\BIBentrySTDinterwordspacing

\bibitem{UCI}
\BIBentryALTinterwordspacing
D.~Dua and C.~Graff, ``{UCI} machine learning repository,'' 2017. [Online].
  Available: \url{http://archive.ics.uci.edu/ml}
\BIBentrySTDinterwordspacing

\bibitem{scikit-learn}
F.~Pedregosa, G.~Varoquaux, A.~Gramfort, V.~Michel, B.~Thirion, O.~Grisel,
  M.~Blondel, P.~Prettenhofer, R.~Weiss, V.~Dubourg, J.~Vanderplas, A.~Passos,
  D.~Cournapeau, M.~Brucher, M.~Perrot, and E.~Duchesnay, ``Scikit-learn:
  Machine learning in {P}ython,'' \emph{Journal of Machine Learning Research},
  vol.~12, pp. 2825--2830, 2011.

\bibitem{Weinberger09}
K.~Q. Weinberger and L.~K. Saul, ``Distance metric learning for large margin
  nearest neighbor classification,'' \emph{The Journal of Machine Learning
  Research}, vol.~10, pp. 207--244, 2009.

\bibitem{metric-learn}
W.~{de Vazelhes}, C.~{Carey}, Y.~{Tang}, N.~{Vauquier}, and A.~{Bellet},
  ``metric-learn: {M}etric {L}earning {A}lgorithms in {P}ython,'' \emph{Journal
  of Machine Learning Research}, vol.~21, no. 138, pp. 1--6, 2020.

\end{thebibliography}

\end{document}